\def\year{2022}\relax
\documentclass[letterpaper]{article} 
\usepackage{aaai22}  
\usepackage{times}  
\usepackage{helvet}  
\usepackage{courier}  
\usepackage[hyphens]{url}  
\usepackage{graphicx} 
\usepackage{natbib}  
\usepackage{caption} 
\DeclareCaptionStyle{ruled}{labelfont=normalfont,labelsep=colon,strut=off} 
\usepackage[ruled, lined,longend]{algorithm2e}
\usepackage{tabularx} 
\usepackage{adjustbox}
\title{Cluster Analysis with Deep Embeddings and Contrastive Learning}
\author{
    Ramakrishnan Sundareswaran, \textsuperscript{\rm 1}
    Jansel Herrera-Gerena, \textsuperscript{\rm 1}
    John Just, \textsuperscript{\rm 1,2}
    Ali Jannesari \textsuperscript{\rm 1}
}
\affiliations{

        \textsuperscript{\rm 1}  Iowa State University\\
    \textsuperscript{\rm 2}  John Deere\\
    ramkris@iastate.edu, janselh@iastate.edu,
    justjohnp@johndeere.com, 
    jannesari@iastate.edu 

}

\usepackage{bibentry}

\begin{document}
\maketitle
\begin{abstract}
Unsupervised disentangled representation learning is a long-standing problem in computer vision. This work proposes a novel framework for performing image clustering from deep embeddings by combining instance-level contrastive learning with a deep embedding based cluster center predictor. Our approach jointly learns representations and predicts cluster centers in an end-to-end manner. This is accomplished via a three-pronged approach that combines a clustering loss, an instance-wise contrastive loss, and an anchor loss. Our fundamental intuition is that using an ensemble loss that incorporates instance-level features and a clustering procedure focusing on semantic similarity reinforces learning better representations in the latent space. We observe that our method performs exceptionally well on popular vision datasets when evaluated using standard clustering metrics such as Normalized Mutual Information (NMI), in addition to producing geometrically well-separated cluster embeddings as defined by the Euclidean distance. Our framework performs on par with widely accepted clustering methods and outperforms the state-of-the-art contrastive learning method on the CIFAR-10 dataset with an NMI score of 0.772, a 7-8\% improvement on the strong baseline.
\end{abstract}
\section{Introduction}
\noindent 
Cluster analysis is one of the fundamental challenges in unsupervised machine learning and has been widely studied. In the context of computer vision, image clustering aims to group data without the presence of ground truth labels.
While this task has been approached from different perspectives, data grouping is generally done based on some well-understood metric for measuring similarity, such as the pair-wise distance between separated clusters. For example, the well-known k-means algorithm \cite{Macqueen67somemethods} assigns each data point to the cluster with the minimum mean based on Euclidean distance and iteratively refines this mean in the feature space. Although this works well for lower dimensional data, this doesn't scale well, and the choice of the feature space becomes more decisive. \\ \\     
To solve this problem of the curse of dimensionality, representation learning using deep neural networks have been widely popularized. Recent methods \cite{DBLP:journals/corr/XieGF15, DBLP:journals/corr/jiangZTTZ16, DBLP:conf/iccv/DizajiHDCH17, Caron_2018_ECCV, shaham2018spectralnet}  use deep neural networks to define a parametrized non-linear mapping from the data space \textit{X} to a lower-dimensional feature space \textit{Z} while optimizing a clustering objective in this representation space. Although optimizing in the representation space yields better clustering, there are still limitations such as poor clustering accuracy, dependency on offline training, requirements for multi-stage training pipelines, and significant dependence on the inherent overlap across classes within a dataset. \\ \\
On the other hand, instance-level contrastive learning has gained a lot of recognition in the self-supervised learning community owing to its recent successes \cite{DBLP:journals/corr/abs-1805-01978,NEURIPS2019_ddf35421,pmlr-v119-chen20j}.  Contrastive learning aims to learn representations such that similar samples stay close to each other while dissimilar ones remain far apart. Most approaches produce different data augmentations in parallel for every instance and aim to learn representations such that augmented samples generate similar embeddings. A contrastive loss in the latent space penalizes and maximizes agreement between differently augmented image embeddings. The ability to push similar instances closer while keeping different samples apart is handy for grouping data and thus performing clustering.  \\ \\
We take inspiration from contrastive representation learning and deep cluster embeddings. Since we need to learn the cluster centers while learning better representations simultaneously, the optimization process has its challenges. To this end, we present a clustering framework that jointly learns representations and predicts cluster centers in an end-to-end manner. Our method consists of a backbone network, an instance-level contrastive head, and cluster centers as learnable parameters. We iteratively refine the cluster centers using an auxiliary target distribution derived from the existing soft probability distribution. Through rigorous ablations studies, we observe that this is essential in addition to the instance-wise contrastive loss for performing single-stage, end-to-end, and online clustering, subsequently generating better outcomes. The main contributions of this work are summarized as follows: (1)  We propose a novel architecture that performs end-to-end online clustering. (2) Define an ensemble loss based on instance-level contrastive loss, KL-divergence based clustering loss, and an aggregated anchor objective function combining the raw cluster centers predicted along with the augmented predictions (3) Conduct extensive experiments and visualizations that present the effectiveness of our framework when compared to widely accepted existing works.
(4) Conduct ablation studies to evaluate the effectiveness of data augmentation, the network's three components, and the anchor objective function.
\begin{figure*}[t]
\centering
\includegraphics[width=0.9\textwidth]{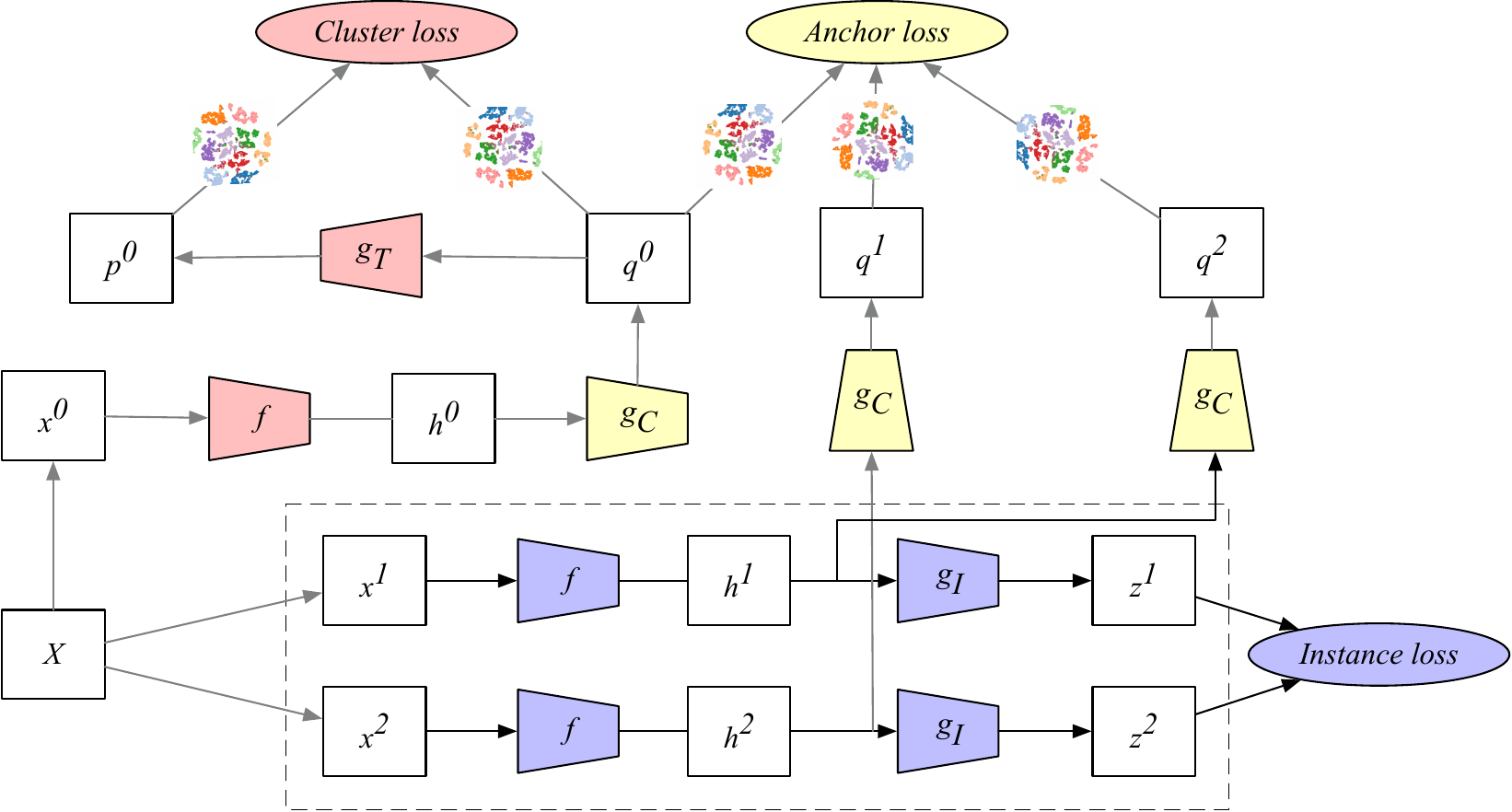} 
\caption{Model Architecture: Two augmented views of one image are processed by $f$(backbone) followed by $g_I$(instance-wise contrastive head) for computing the Instance loss. A raw input sample is fed to $f$(backbone) followed by $g_C$ and $g_T$ for computing the Cluster loss. A combination of the raw input representations and augmented representations($h^0, h^1$ and $h^2$) is fed to $g_C$ for computing the anchor loss. The net loss is computed using the Instance loss, Cluster loss and Anchor loss.}
\label{fig1}
\end{figure*}
\section{Related Work}
\textbf{Contrastive Learning:} Self-supervised representation learning has seen substantial growth in recent years, with huge successes in extending these learned representations for solving downstream tasks. While traditional methods \cite{NIPS2014_07563a3f, DBLP:journals/corr/NorooziF16, DBLP:journals/corr/abs-1803-07728}  aim to learn representations by solving pretext tasks, the advent of SimCLR \cite{pmlr-v119-chen20j} popularized contrastive learning methods, which are not limited to noncomprehensive representations learned while solving specific pretext tasks. Several other works \cite{DBLP:journals/corr/abs-1805-01978,DBLP:journals/corr/abs-1807-03748,DBLP:journals/corr/abs-1905-09272,DBLP:journals/corr/abs-1906-05849,hjelm2019learning,DBLP:journals/corr/abs-1911-05722} prior to SimCLR also used certain elements of contrastive learning to achieve reasonable success. The overarching theme in all these works is that a set of sub-samples that are drawn from the same instance of data (positive pair) are optimized to be closer in the embedding space, while samples from other data instances (negative pair) get pushed apart. Due to the absence of annotated labels, a positive pair often consists of data augmentations of the sample, whereas negative pairs consist of the sample and randomly chosen samples from the minibatch. Therefore, different instances are well-separated in the learned embedding space while actively preserving local invariance for each instance.  \\ \\ While heavy data augmentation, large batch size, and hard negative mining remain the key components for the success of contrastive methods by avoiding collapse to trivial solutions, more recent works have highlighted other interesting viewpoints and translated this approach to other domains. For example, BYOL \cite{DBLP:journals/corr/abs-2006-07733} achieves state-of-the-art results without using negative pairs; CURL \cite{DBLP:journals/corr/abs-2004-04136} applies ideas from contrastive learning to reinforcement learning; SimSiam \cite{Chen2021ExploringSS} uses a stop-gradient operation without using negative pairs, large batches, and momentum encoders; GraphCL \cite{DBLP:journals/corr/abs-2010-13902} proposes a contrastive framework for learning unsupervised representations on graph data; SwAV \cite{DBLP:conf/nips/CaronMMGBJ20} takes advantage of contrastive methods without requiring to compute pairwise comparisons by simultaneously clustering data while enforcing consistency between cluster assignments produced for different views of the same image; Barlow Twins \cite{DBLP:journals/corr/abs-2103-03230} feeds two distorted versions of samples into the same network to extract features and learns to make the cross-correlation matrix between these two groups of output features close to the identity. \\ \\ In the same timeline, some works \cite{DBLP:journals/corr/abs-2004-11362, DBLP:journals/corr/abs-2007-13916, DBLP:journals/corr/abs-2005-04966} show that while instance-wise contrastive learning groups similar instances together \cite{DBLP:journals/corr/abs-1805-01978} by pushing representations from different instances apart, it gives rise to worse representations in some cases because of overlooking the inherent semantic similarity in the dataset. 
\\ \\
\textbf{Unsupervised Deep Clustering:} Traditional clustering algorithms such as K-Means \cite{Macqueen67somemethods} and Gaussian Mixture Models \cite{CELEUX1995781} do not perform well with larger datasets due to their inability to learn representations in higher dimensions \cite{Steinbach03thechallenges}. This drawback led to methods such as DEC\cite{DBLP:journals/corr/XieGF15} that use neural networks to perform efficient clustering. While DEC learns a mapping from the data space to a lower-dimensional feature space where it iteratively optimizes a clustering objective, JULE\cite{DBLP:journals/corr/YangPB16} follows a recurrent process where it progressively merges data points and uses the clustering results as a supervisory signal to learn a more discriminative representation by a neural network. Although works like DEC do not surpass modern methods in terms of clustering performance, the ideas introduced in this work are inspiring. Similar ideas have also motivated successes  in other domains, such as text clustering\cite{hadifar-etal-2019-self,DBLP:journals/corr/abs-2103-12953}. Likewise, DeepCluster \cite{Caron_2018_ECCV} groups the features using k-means and trains a convolutional neural network with the cluster indices for representation learning.  However, a significant drawback with such two-stage methods is that the performance deteriorates due to errors accumulated during the alternation. Moreover, the successes of these methods in online scenarios are limited as they need to perform clustering on the entire dataset. \\ \\ Subsequently, there have been more recent works that perform end-to-end clustering and achieve state-of-the-art results. For example, IIC \cite{DBLP:journals/corr/abs-1807-06653} maximizes the mutual information between the class assignments of two different views of the same image to learn representations that preserve what is shared between the views while discarding instance-specific details. PICA \cite{Huang_2020_CVPR} learns the most semantically plausible data separation of the clustering solution by performing partition confidence maximization. \\ \\In a more recent approach, Contrastive Clustering(CC) \cite{Li_Hu_Liu_Peng_Zhou_Peng_2021} performs instance and cluster level contrastive learning in the row and column space, respectively, by maximizing the similarities of positive pairs while minimizing those of negative ones. This work uses a SimCLR \cite{pmlr-v119-chen20j} like loss and incorporates an entropy term to avoid collapse to a trivial solution. Inspired by similar ideas, our work builds upon instance-wise contrastive learning along with additional components, creating an ensemble learning algorithm for deep clustering. Our method considers CC as the baseline and outperforms it on all datasets considered. \\ \\ Analogously, IDFD \cite{DBLP:journals/corr/abs-2106-00131} combines instance discrimination and softmax-formulated decorrelation constraints, simultaneously learning representations and clustering, and ConCURL \cite{DBLP:journals/corr/abs-2105-01289} performs deep clustering by building a consensus on multiple clustering outputs, both yielding state-of-the-art results. Besides, there have been recent works \cite{Gansbeke2020SCANLT,DBLP:journals/corr/abs-2103-09382} that do not perform end-to-end clustering. For example, SCAN \cite{Gansbeke2020SCANLT} uses embedding features of the representation learning model for computing the instance similarity, consequently using the learned representations for finding the closest images to an anchor image using the nearest samples; SPICE \cite{DBLP:journals/corr/abs-2103-09382} generates pseudo-labels via self-learning and directly uses the pseudo-label-based classification loss to train a deep clustering network, which requires a fair amount of supervised pre-training of parameters. 
\section{Model}
As illustrated in Figure \ref{fig1}, our method consists of three jointly learned components along with their respective losses. These components include an instance-wise contrastive head, a clustering head, and an anchor head. Additionally, we employ a backbone network that generates a mapping to a lower-dimensional representation space, followed by the corresponding components, thus computing the contrastive, clustering, and anchor losses. Furthermore, our framework utilizes augmented data in addition to the raw images, which are mainly applicable for computing the contrastive and anchor losses.  In this section, we will explain the details of each of these components, an instance-wise contrastive head, a clustering network, an anchor network, and ultimately define the ensemble loss used for training the network end-to-end.
\subsection{Backbone} We choose the commonly used ResNet34 \cite{DBLP:journals/corr/HeZRS15} as the backbone for our network, denoted by $f$($\cdot$), owing to its success in contrastive learning and for a fair comparison with our baseline. The neural network $f$($\cdot$) extracts representations in a lower-dimensional space for the input samples. Besides, we use a data augmentation module that produces two correlated views, denoted by $x_1$ and $x_2$ for every input sample $x$. Since data augmentation plays a critical role \cite{pmlr-v119-chen20j} in achieving good performance for downstream tasks, we use Resized Crop,
Color Jitter, Grayscale, Horizontal Flip, and Gaussian Blur similar to \cite{Li_Hu_Liu_Peng_Zhou_Peng_2021} and apply them stochastically and independently with the same settings as SimCLR \cite{pmlr-v119-chen20j}. We pass the augmented and original input samples through the backbone network $f$($\cdot$), generating
$f(x\textsubscript{1}) = h\textsubscript{1}$, $f(x\textsubscript{2}) = h\textsubscript{2}$, and $f(x\textsubscript{0}) = h\textsubscript{0}$, respectively. While the anchor head requires the augmented as well as raw input samples, the instance-wise contrastive head and clustering head require only $h\textsubscript{1}$ and $h\textsubscript{2}$, and $h\textsubscript{0}$, respectively. 
\subsection{Instance-wise Contrastive Head}
We choose a neural network $g\textsubscript{I}$($\cdot$) that maps representations to the latent space where the instance-wise contrastive loss gets applied. We use the representations of the augmented data samples as the input to the instance-wise contrastive head, namely, $h_1$ and $h_2$. This contrastive head aims to maximize the similarity of positive pairs while penalizing the negative ones. Formally, given a minibatch $B$ of size $N$, we generate two data augmentations resulting in a minibatch $B^a$ with $2N$ data samples $\{ x^1_1,x^1_2,...,x^1_N,x^2_1,x^2_2,...x^2_N\} $. Let $x^1_k$ belongs to $B^a$, be the first augmentation of the $k^t{}^h$ input sample in $B$, and $x^2_k$, be the second augmentation of the $k^t{}^h$ input sample from $B$. We consider $\{ x^1_k,x^2_k\} $ as a positive pair in the minibatch and the remaining $2N-2$ examples as negative pairs. For the $k^t{}^h$ sample in minibatch $B^a$, the backbone layer generates representations for a positive pair $\{ h^1_k,h^2_k\} $ along with $2N-2$ negative pairs. Thus, the contrastive head ${g\textsubscript{I}(.)}$ takes these representations as inputs producing
$z^1_k = g_I(h^1_k)$.  We choose a two-layer MLP, with one hidden layer, thus reducing the induced information loss by projecting the feature matrix into a subspace via $z^1_k = g_I(h^1_k) = W^{(2)}\sigma(W^{(1)}h^1_k)$ where $\sigma$ denotes the ReLU non-linearity. Thus, without loss of generality, for a sample $x^1_i$, we apply the following loss: 
\[ l^I_{i^1} = -log\frac{exp(\psi(z^1_i,z^2_i)/\tau)}{\sum_{j=1}^{N} [exp(\psi(z^1_i,z^1_j)/\tau) + exp(\psi(z^1_i,z^2_j)/\tau)]} \] where  $\tau$ denotes the instance-wise contrastive learning temperature parameter which is set to 0.5 and $\psi(x,y)$ computes the pairwise similarity computed measure by the cosine distance, i.e,
\[ \psi(z^1_i,z^2_i) = \frac{(z^1_i) \cdot (z^2_i)^T}{\| z^1_i \| \| z^2_i\| } \]
To compute the net instance loss, we compute $l^I_{i^1}$ for all the augmented samples in the minibatch $B^a$,
\[ {\mathcal{L}}_{Instance} = \sum_{i=1}^{N} (l^I_{i^1} + l^I_{i^2})/2N \]
\newcommand{\ra}[1]{\renewcommand{\arraystretch}{#1}}
\begin{algorithm}[ht!]
    \SetKwFunction{isOddNumber}{isOddNumber}
    \SetKwInOut{KwIn}{Input}
    \SetKwInOut{KwOut}{Output}
    \KwIn{dataset $X$; training epochs $E$; batch size $N$; temperature parameter $\tau$ ; $M$ clusters; structure of $T , f, g_I. g_C, g_T$}
    \KwOut{cluster assignments}
    \tcp{Train}
    \For{$epoch = 1$ \KwTo $E$}{
         \For{$i \in \{1, \ldots, N\}$}{
            sample a minibatch B\;
            \tcc{store raw sample}
            $x^0_i = x_i$\;
            sample two augmentations $T^1 \sim \mathcal{T}, T^{2} \sim \mathcal{T}$\;
            \tcc{generate augmentations}
            $x^1_i = T^1(x_i)$\;
            $x^2_i = T^2(x_i)$\;
            generate representations as follows\;
            \tcc{instance}
            $h^1_i = f(x^1_i), h^2_i = f(x^2_i)$\;
            $z^1_i = g_I(h^1_i), z^2_i = g_I(h^2_i)$\;
            \tcc{cluster}
            $h^0_i = f(x^0_i)$\;
            $q^0_i = g_C(h^0_i), p^0_i = g_T(q^0_i)$\;
            \tcc{anchor}
            $h^0_i = f(x^0_i), h^1_i = f(x^1_i), h^2_i = f(x^2_i)$\;
            $q^0_i = g_C(h^0_i), q^1_i = g_C(h^1_i), q^2_i = g_C(h^2_i)$\;
            compute $\mathcal{L}_{Instance}, \mathcal{L}_{Cluster}$ and $  \mathcal{L}_{Anchor}$\;
            compute overall loss $\mathcal{L}_{Total}$\;
            update $f,g_I,g_C$ to minimize $\mathcal{L}_{Total}$\;
            
        }
    }
    \tcp{Test}
    \For{$x$ in $X$}{
        $h = f(x)$ \tcp*{generate embeddings}
        $p = g_C(h)$ \tcp*{cluster probability}
        assign $x$ to cluster $c$ using $p$\;
    }
    \caption{}
\end{algorithm}
\subsection{Clustering network}
While the contrastive head brings similar instances closer, the clustering head maximizes agreement between semantically closer samples, thereby bringing samples that cluster together, resulting in compact clusters. Given K initial cluster centroids $\{ \mu \} _{j=1}^{k}$ computed using K-Means, we perform a two-phase parameter optimization. In the first phase, we compute soft cluster assignments by using the Student's t-distribution\cite{JMLR:v9:vandermaaten08a} as a kernel to measure similarity between embedding $h_i$ and centroid $\mu_j$: \[ q_{ij} = \frac{(1 + \| h_i - \mu_{j^{'}} \|^{2}/\alpha)^{-\frac{(\alpha+1)}{2}}}{\sum_{j^{'}} (1 + \| h_i - \mu_{j^{'}} \|^{2}/\alpha)^{-\frac{(\alpha+1)}{2}}} \]
where $h_i = f(x_i)$ where $x_i\in B$, which denotes a minibatch with unaugmented samples, $\alpha$ are the degrees of freedom, and $q_{ij}$ is the probability of assigning sample i to cluster j. We set $\alpha=1$ for all our experiments as per \cite{JMLR:v9:vandermaaten08a}. In the second phase, we iteratively refine the cluster centers by learning from  high confidence assignments using an auxiliary target distributed inspired by \cite{DBLP:journals/corr/XieGF15}. Specifically, the auxiliary target distribution $p_{ij}$ is defined as follows: \[ p_{ij} = \frac{q^{2}_{ij}/f_j}{\sum_{j^{'}}q^2_{ij}/f_{j^{'}}} \] where $f_j = \sum_{j}q_{ij} $ are soft cluster frequencies approximated within a minibatch $B$. This target distribution fosters learning with high confidence by first raising $q_{ij}$ to the second power, resulting in finer soft probabilities and then normalizing by the cluster frequency, handling any implicit bias introduced due to imbalanced clusters. 
\begin{table*}[ht!]
    \begin{adjustbox}{width=\textwidth}
  \begin{tabular}{lllllllllllll}\\
    \hline
     Dataset
    &\multicolumn{3}{c}{\textbf{CIFAR-10}}
    &\multicolumn{3}{c}{\textbf{ImageNet-10}}
    &\multicolumn{3}{c}{\textbf{ImageNet-Dogs}}
    &\multicolumn{3}{c}{\textbf{Tiny- ImageNet}}\\
    \cline{1-13} 
    Metric &  NMI & ACC & ARI & NMI & ACC & ARI & NMI & ACC & ARI & NMI & ACC & ARI \\ 
   \hline
   K-Means & 0.087 & 0.229 & 0.049 & 0.119 & 0.241 & 0.057 & 0.055 & 0.105 & 0.020 & 0.065 & 0.025 & 0.005 \\
   SC & 0.103 &  0.247 &  0.085 &  0.151  & 0.274 &  0.076 &  0.038  & 0.111  & 0.013 &  0.063 &  0.022 &  0.004\\
   AE &  0.239 &  0.314 &  0.169 & 0.210 & 0.317 & 0.152 & 0.104 & 0.185 & 0.073 & 0.131 & 0.041 & 0.007\\
   DAE  & 0.251 &  0.297 &  0.163 &  0.206 &  0.304 &  0.138 &  0.104 &  0.190 &  0.078  & 0.127  & 0.039 &  0.007\\
   DCGAN  &  0.265  &  0.315 &   0.176 & 0.225 &   0.346 &   0.157 &   0.121 &   0.174 &   0.078 &   0.135 &   0.041 &   0.007\\
   VAE   &  0.245   &  0.291  &   0.167  & 0.193   &  0.334  &   0.168  &   0.107   &  0.179  &   0.079   &  0.113   &  0.036  &   0.006\\
   JULE  &   0.192   &  0.272  &   0.138   & 0.175   &  0.300  &   0.138  &   0.054   &  0.138   &  0.028   &  0.102   &  0.033  &   0.006\\
   DEC   &   0.257   &   0.301   &   0.161   &  0.282    &  0.381   &   0.203    &  0.122    &  0.195   &   0.079    &  0.115   &   0.037    &  0.007\\
   DAC    &  0.396   &   0.522    &  0.306    &   0.394   &   0.527   &   0.302   &   0.219   &   0.275   &   0.111   &   0.190   &   0.066   &   0.017\\
   ADC & -  & 0.325 &  - &   -  & - &  - &  -  & -  & -  & -  & -  &  -\\
   DDC & 0.424 &  0.524 &  0.329 &  0.433  & 0.577  & 0.345  & - &  -  & -  & -  & -  & -  \\
   DCCM  & 0.496 &  0.623 &  0.408  & 0.608  & 0.710 &  0.555 &  0.321  & 0.383 &  0.182  & 0.224  & 0.108 &  0.038\\
      PICA    &  0.591    &  0.696   &   0.512   & 0.802   &   0.870   &   0.761   &   0.352   &   0.352    &  0.201   &   0.277    &  0.098    &  0.040\\
     \hline
  CC(Baseline) &  0.705 &  0.790  & 0.637 &  0.859 &  0.893 &  0.822  & 0.445  & 0.429 &  0.274 &  0.340  & 0.140  & \textbf{0.071}\\
  Ours(Representation) &  0.766 & 0.805 & 0.703 & 0.776 & 0.598 & 0.536 & 0.525 & 0.534 & 0.275 & \textbf{0.346} & \textbf{0.158} & 0.069 \\
Ours(Cluster) &  \textbf{0.772} & 0.805 & 0.698 & 0.804 & 0.674 & 0.588 & 0.326 & 0.281 & 0.069 & 0.292 & 0.100 & 0.042 \\
   \hline
   More recent methods \\
   \hline
       Single-Noun Prior & 0.731 & \textbf{0.853} & 0.702 & - & - & -  & 0.505 & 0.551 & 0.381 & - & - & - \\
   ConCURL & 0.762 & 0.846 & \textbf{0.715} & \textbf{0.907} & \textbf{0.958} & \textbf{0.909} & \textbf{0.63} & \textbf{0.695} & \textbf{0.531} & - & - & -\\
   IDFD & 0.711 & 0.815 & 0.663 & 0.898 & 0.954 & 0.901 & 0.546 & 0.591 & 0.413 & - & - & -\\
   \hline
 \end{tabular}
 \end{adjustbox}
   \caption{Comparative study on four challenging image datasets}
  \label{Comparison}
\end{table*}
Consequently, we define our clustering objective as a KL divergence loss between the cluster assignments $q_{i}$ and target distribution $p_{i}$ as follows: \[ l^C_i = \mbox{KL}[p_i \| q_i] = \sum_{i}\sum_{j} p_{ij}log\frac{p_{ij}}{q_{ij}} \] To compute the net clustering loss, we compute $l^C_i$ for all the samples in minibatch B, 
\[ \mathcal{L}_{Cluster} = \sum_{i=1}^{N} l^C_{i}/N \]
\subsection{Anchor network}
While the clustering loss brings semantically closer clusters together, it only performs parameter optimization over a minibatch B that consists of raw image samples. We propose an anchor loss such that cluster assignments for raw samples explicitly maximize agreement with augmented cluster assignments using an aggregated KL divergence loss. Let the anchor distribution for the $i^{th}$ sample be $q^0_{i}$, and $q^1_{i}$ and $q^2_{i}$ denote the augmented cluster assignments. While the Jenson-Shannon divergence(JSD) produces a symmetrized and smoothed version of KL divergence by considering the mixture distribution of two probability distributions, we pair an anchor distribution with two probability distributions that smoothens the combined KL divergence loss without changing the structure and symmetry of the objective function. More specifically, we compute the loss as follows:
\[ l^A_i = \mbox{KL}[q^0_i  \|  q^1_i] +  \mbox{KL}[q^0_i \| q^2_i] \]
To compute the net anchor loss, we compute $l^A_i$ for all the samples in minibatch B, 
\[ \mathcal{L}_{Anchor} = \sum_{i=1}^{N} l^A_{i}/N \]
\subsection{Objective function}
Optimizing the instance-wise contrastive head, clustering network, and the anchor network is one-state and end-to-end learning. In summary, the overall objective is the cumulative sum of the contrastive loss, clustering loss, and anchor loss as follows:
\[ \mathcal{L}_{Total} = \alpha\mathcal{L}_{Instance} + \beta\mathcal{L}_{Clustering} + \gamma\mathcal{L}_{Anchor} \]
where $\alpha,\beta$ and $\gamma$ denote the weight parameters to balance out losses for better performance \cite{NEURIPS2020_f3ada80d}.  Although we don't conduct an extensive tuning of these hyperparameters, we observe that setting $\alpha=20 ,\beta=0.1$, and $\gamma=0.1$ prioritizes instance-wise contrastive learning early on and fine-tunes the clusters with the clustering and anchor loss, yielding reasonably high accuracy. Having a higher weight for contrastive learning loss helps in learning effective representations and influences the movement of the gradient towards the spectrum of the global optimum. In contrast, the clustering and anchor weights help fine-tune the cluster centers once the gradient is within the narrow horizon for achieving good clustering. Another viable option is considering a more complicated learning schedule with a drop in alpha as time progresses, while values of beta and gamma increase. However, we omit this in our work since the existing more straightforward schedule performs well. 
\section{Results}
\subsection{Datasets}
We evaluate our proposed framework on four challenging image datasets. Table \ref{Datasets} summarizes each of these datasets. 
\begin{table}[ht!]
\begin{tabularx}{0.45\textwidth}{|c|c|c|X|}
\hline
  \textbf{Dataset} & \textbf{Split}  & \textbf{Samples}  & \textbf{Classes} \\
\hline
CIFAR-10 & Train+Test & 60K & 10\\
ImageNet-10 & Train & 13K & 10\\
ImageNet-Dogs & Train & 19.5K & 15\\
Tiny-ImageNet & Train & 100K & 200 \\
\hline
\end{tabularx}
  \caption{Summary of datasets}\label{Datasets}
\end{table}
\begin{figure*}[t!]
\centering
\includegraphics[width=\textwidth]{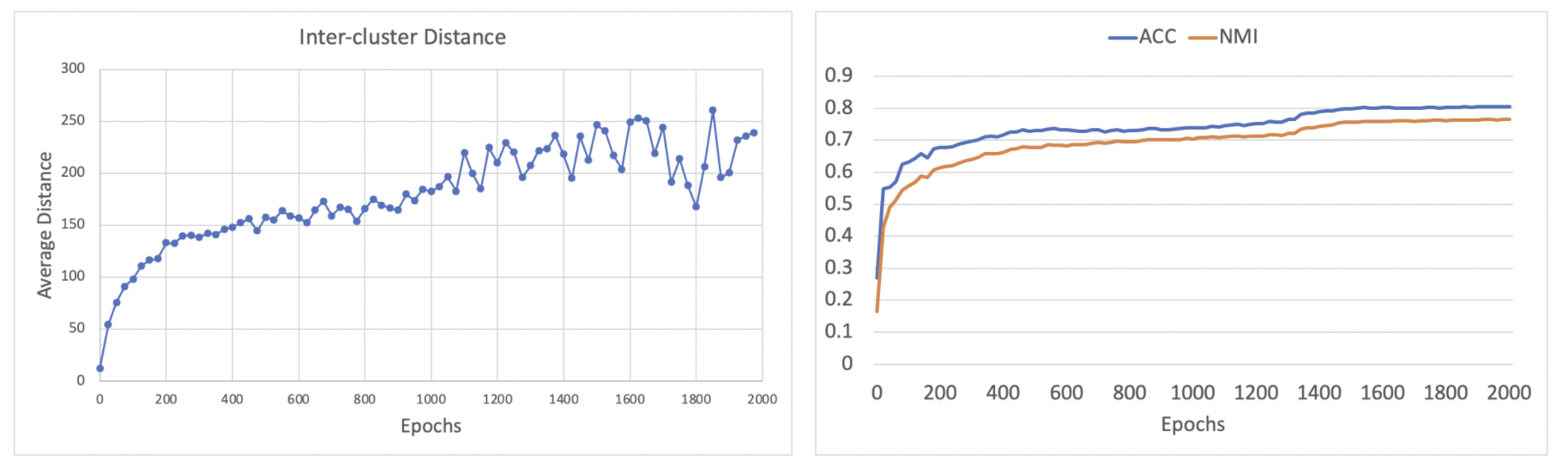} 
\caption{Evolution of inter-cluster distance, Accuracy and NMI scores with epochs on the CIFAR-10 dataset}
\label{fig2}
\end{figure*}
\begin{figure*}[ht!]
\centering
\includegraphics[width=0.9\textwidth]{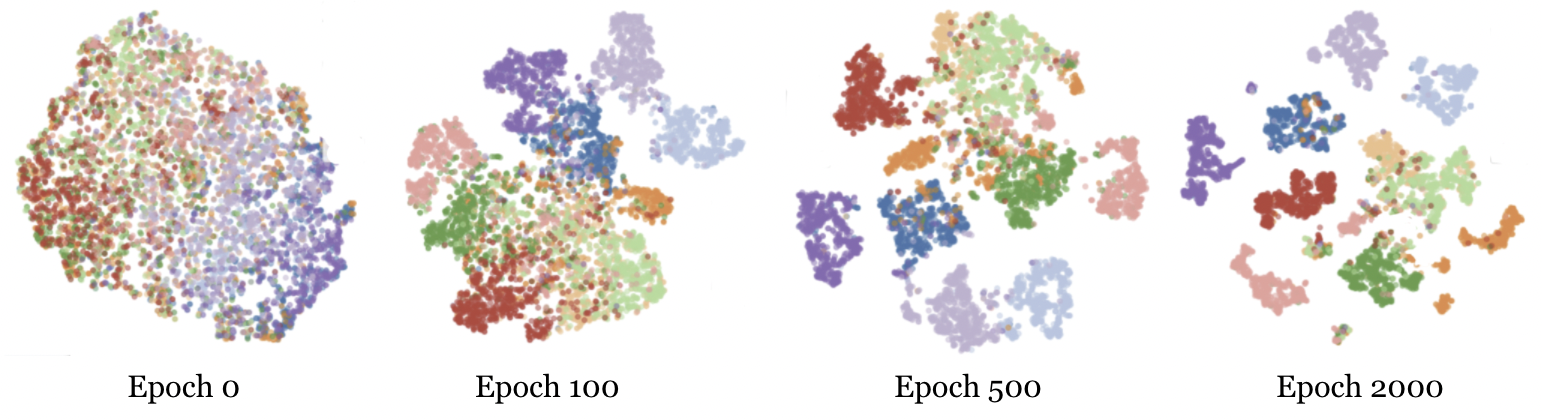} 
\caption{Evolution of learned representations on the CIFAR-10 datset. The t-SNE features are based on the reprsentations learned by the Instance-wise contrastive head and the colors denote the class predicted by the network.}
\label{fig3}
\end{figure*}
While we use both the training and testing data for CIFAR-10, we only use the training set for ImageNet-10, ImageNet-Dogs, and Tiny-ImageNet.
\subsection{Implementation Details}
We implement our model using PyTorch \cite{NEURIPS2019_9015}. We choose ResNet34 as our network backbone for a fair comparison with our baseline. Besides, we resize all images to $224\mbox{ x }224$, followed by respective image transformations for generating augmented data. We omit Gaussian Blur augmentation for CIFAR-10 and Tiny ImageNet since datasets with small image samples tend to get blurred. Thus, our minibatch generator yields a minibatch with a set of raw image samples as well as a pair of augmented data samples. For the instance-wise contrastive head, we use a linear layer that takes input features of $512$ dimensions produced by the backbone and outputs vectors of size $128$ dimensions, performing a mapping to lower-dimensional latent space for computing the contrastive loss.  We choose the instance-level temperature parameter as $\tau=0.5$ for optimizing the contrastive loss. For the clustering head, we initialize a linear network with learnable parameters depending on the number of clusters; thus, our clustering head is of size $512 \mbox{ x } K$, where $512$ features are output from the backbone and K indicates the number of clusters. We choose $\alpha=1$ for all datasets similar to \cite{DBLP:journals/corr/XieGF15}.  We use Adam\cite{DBLP:journals/corr/KingmaB14} as the optimizer with a learning rate of $0.0003$ without any weight decay for the entire model. Due to memory limitations, the batch size is set to 128; however, we train the model longer (2000 epochs) to compensate for the smaller batch size \cite{pmlr-v119-chen20j}. We conduct our experiments using Nvidia Tesla V100 32G and it takes around 70 hours to train our model for 1000 epochs on the CIFAR-10 dataset.
\subsubsection{Evaluation Metrics:}For performing a comparative assessment, we consider three metrics, namely, Normalized Mutual Information(NMI), Accuracy(ACC), and Adjusted Rand Index(ARI), for evaluating our method. For these metrics, a higher value indicates better clustering. We report these metrics on the cluster centers predicted as well as on the learned representations by extracting features from the instance-wise contrastive head followed by K-Means.
\subsection{Comparison with State-of-the-art}
We demonstrate that our framework can achieve highly competitive performance on image clustering by evaluating on four challenging image datasets and performing a comparative study. For benchmarking, we consider state-of-the-art clustering approaches, including K-Means (\citeauthor{Macqueen67somemethods}),
SC (\citeauthor{Zelnik-manor04self-tuningspectral}), AE (\citeauthor{10.5555/2976456.2976476}), DAE (\citeauthor{JMLR:v11:vincent10a}), DCGAN (\citeauthor{DBLP:journals/corr/RadfordMC15}), VAE (\citeauthor{DBLP:journals/corr/abs-1906-02691}), JULE (\citeauthor{DBLP:journals/corr/YangPB16}), DEC (\citeauthor{DBLP:journals/corr/XieGF15}), DAC (\citeauthor{Chang_2017_ICCV}),
ADC (\citeauthor{10.1007/978-3-030-12939-2_2}), DDC (\citeauthor{DBLP:journals/corr/abs-1905-01681}), DCCM (\citeauthor{Wu_2019_ICCV}), PICA(\citeauthor{Huang_2020_CVPR}), CC (\citeauthor{Li_Hu_Liu_Peng_Zhou_Peng_2021}) Single-Noun Prior (\citeauthor{DBLP:journals/corr/abs-2104-03952}), ConCURL(\citeauthor{DBLP:journals/corr/abs-2105-01289}), and IDFD (\citeauthor{DBLP:journals/corr/abs-2106-00131}). As shown in Table \ref{Comparison}, our model, either the representations learned followed by K-Means or the clusters predicted, outperforms the chosen baseline(CC) across all datasets on all metrics. Moreover, there is almost a 7-8\% increase in the NMI score on the CIFAR-10 dataset, which is substantial given the already high NMI score achieved using CC. As shown in Figure 2, along with NMI and Accuracy increasing, the inter-cluster distance also increases with time, proving that our model converges, producing well-separated clusters. Besides, we observe that while the model representation gives higher scores on the ImageNet-Dogs and Tiny-ImageNet datasets, the end-to-end clusters predicted fare better on the CIFAR-10 and ImageNet-10 datasets. While the model is robust, we suppose that a more thorough hyperparameter optimization would allow better clustering scores over the representations learned for different datasets since the number of classes and the inherent structure of the data have a profound impact on the gradients and the net ensemble loss. We also observe that our framework has comparable performance with scores reported in more recent approaches such as ConCURL, Single-Noun Prior, and IDFD. Each method is a better performer across various metrics and different datasets. 
\subsection{Qualitative Study}
While quantitative metrics such as NMI and Accuracy, combined with the increase in the inter-cluster distance over time, show the efficacy of our framework, we compute the t-SNE plots over time to evaluate the learned representations. We determine these embeddings using the output of the instance-wise contrastive head and perform cluster assignments using the predicted cluster centers. The results in figure \ref{fig3} show that our model converges well since the cluster assignments become more distinct and separated over time, transitioning from one big block to scattered features grouped together based on labels. 
\subsection{Ablation Study}
\subsubsection{Importance of Data  Augmentations:} We reaffirm the importance of data augmentations  \cite{pmlr-v119-chen20j} by observing
\begin{table}[ht!]
\centering
\begin{tabular}{||c c c c||} 
 \hline
 Input($x_0,x_1,x_2$) & NMI & ACC & ARI \\ [0.5ex] 
 \hline
 $x + x + x$  & 0.038 & 0.163  & 0.197 \\ 
 $T(x) + T(x) + T(x)$ & 0.679 & 0.707 & 0.537 \\
 $x + T(x) + T(x)$ & 0.772 & 0.805 & 0.698  \\
 \hline
\end{tabular}
\caption{Effect of data augmentations}
\label{table:3}
\end{table}that our method relies upon data augmentations for achieving better clustering. In Table \ref{table:3}, $T(x)$ denotes augmented data, and $x_0, x_1, x_2$ are the respective inputs, along with the clustering scores on the CIFAR-10 dataset.We also follow that while augmenting data is critical for instance-wise contrastive learning, it is not necessary, in fact, detrimental, for our clustering network. 
\subsubsection{Effect of respective losses:}
To prove the effectiveness of the three components in our framework, we conduct an ablation study on the CIFAR-10 dataset by discarding certain losses, as shown in Table \ref{table:4}. We use the representation followed by K-means to compute the metrics for the case with only the instance-wise contrastive head. The results suggest that the cluster and anchor loss are essential components. 
 \begin{table}[ht!]
\centering
\begin{tabular}{||c c c c||} 
 \hline
 Loss & NMI & ACC & ARI \\ [0.5ex] 
 \hline
 Instance  & 0.699 & 0.782  & 0.616 \\ 
 Instance + Cluster & 0.729 & 0.764 & 0.630 \\
 Instance + Cluster + Anchor & 0.772 & 0.805 & 0.698  \\
 \hline
\end{tabular}
\caption{Effect of each loss on the CIFAR-10 dataset}
\label{table:4}
\end{table}
\subsubsection{Effect of Anchor Objective function:}While the previous ablation shows the necessity of each component, we conduct another ablation study on the CIFAR-10 dataset with other potential functions for proving the chosen anchor objective's effectiveness. Our experiments show that pushing the anchor assignments to each augmented assignment using an aggregated KL divergence loss gives the best results. 
\begin{table}[ht!]
\centering
\begin{tabular}{||c c c c||} 
 \hline
 Function & NMI & ACC & ARI \\ [0.5ex]
 \hline
 $\mbox{JSD}[q^1_i\|q^2_i]$ & 0.730 & 0.768 & 0.617 \\
  $\mbox{KL}[p^0_i  \|  q^1_i] +  \mbox{KL}[p^0_i \| q^2_i]$ &  0.728 & 0.749 & 0.641 \\
    $\mbox{KL}[p^1_i  \|  q^2_i] +  \mbox{KL}[p^2_i \| q^1_i]$ & 0.718 & 0.651 & 0.578 \\
  $\mbox{KL}[q^0_i  \|  q^1_i] +  \mbox{KL}[q^0_i \| q^2_i]$  & 0.772 & 0.805 & 0.698\\
 \hline
\end{tabular}
\caption{Effect of different Anchor objective functions}
\label{table:5}
\end{table}
\section{Conclusion}
We proposed a new method for image clustering using deep embeddings and contrastive learning, with an ensemble loss that depends on instance-wise contrastive learning, clustering, and an anchor component. This framework can be viewed as an unsupervised technique to learn relevant representations efficiently.  Our empirical evaluation on four datasets shows the ability of our framework to generalize and produce high evaluation scores, with up to 7-8 percentage points NMI gain on the CIFAR-10 dataset. The proposed framework shows that its promising to use multiple synergetic elements that focus on different features, providing scope to learn better representations. For future work, we plan to assist the training using weak supervision, promoting  learning better representations.
\appendix

\bibliography{aaai22}
\end{document}